\documentclass[11pt,a4paper]{article}
\usepackage[hyperref]{acl2020}
\usepackage{times}
\usepackage{latexsym}

\usepackage{microtype}
\usepackage[T1]{fontenc}
\usepackage[utf8]{inputenc}
\inputencoding{utf8}
\usepackage{textcomp}
\usepackage{tipa}
\usepackage{latexsym}
\usepackage{graphicx}

\aclfinalcopy 

\title{Tone prediction and orthographic conversion for Basaa}

\author{Ilya Nikitin \\ HSE University, Russia\\  \texttt{ianikitin93@gmail.com} \\
         \And Brian O'Connor \\ Indiana University Bloomington, USA \\ \texttt{briangerardoconnor@gmail.com} \\\\
         {\bf Anastasia Safonova} \\ HSE University, Russia \\ \texttt{an.saphonova@gmail.com}}

\date{}

\begin{document}
\maketitle
\begin{abstract}
In this paper, we present a seq2seq approach for transliterating missionary Basaa orthographies into the official orthography. Our model uses pre-trained Basaa missionary and official orthography corpora using BERT. Since Basaa is a low-resource language, we have decided to use the mT5 model for our project. Before training our model, we pre-processed our corpora by eliminating one-to-one correspondences between spellings and unifying characters variably containing either one to two characters into single-character form. Our best mT5 model achieved a CER equal to 12.6747 and a WER equal to 40.1012.
\end{abstract}

\section{Introduction}

Basaa is a language spoken in Cameroon by a population of approximately 300,000 people \cite{24}. During Cameroon’s history, there were Catholic and Protestant missionaries who came over and wrote their respective bibles in the Basaa language. The issue was that the missionaries used two different orthographies for Basaa, using different diacritics. Diacritics are distinguishing markings that are placed on letters to indicate a different sound from the letters that are without diacritics. Another issue was the lack of indication for the tones in Basaa, which would be reflected in the official orthography of Basaa by the Academy of Languages of Cameroon.

The purpose of this project is to create a program that can use tone prediction to convert Catholic and Protestant orthographies to the official orthography. This will be aided by the use of BERT and mT5, tools for pre-processing our corpora and creating our language model respectively. For the purpose of our project, we will be using mT5 to help transliterate Catholic and Protestant texts to the official orthography with the use of machine learning. The model will learn the necessary tonal, diacritic, and character differences between the missionary orthographies and the official orthography to be able to more accurately guess the correct spelling of masked tokens.

\subsection{Basaa Phonetics and Phonology}

The Basaa language has 30 consonant sounds. The nasal (including prenasalized stops) and fricative consonants are predominant: they are formed almost at all places of articulation. All voiceless stops occur only at the beginning of a word, and in other positions undergo lenition to fricatives and taps (in intervocalic position it becomes voiced): /p/ goes to /\textphi/ and /\textbeta/, /t/ – to /\d{\textfishhookr}/ and /\textfishhookr/, and /k/ – to /\textchi/ and /\textgamma/. Voiced stops occurs only as part of a prenasalized stop: there are not any /b/, /d/ or /g/, but only /\textsuperscript{m}b/, /\textsuperscript{n}d/ and /\textsuperscript{n}g/. The implosive sound /\texthtb/ is pronounced at the absolute beginning of the word or morpheme if it is preceded by a sonorant. After a vowel, the implosive is realized as a voiced fricative /\textbeta/. There is a free variation of alveolar and glottal fricatives in the pre-pausal position, for example: [m\'{o}m-\'{o}h] ~ [m\'{o}m-\'{o}s] ‘to console’ (Makasso, J. Lee 2015).
There are 7 vowels in Basaa: /i/, /e/, /\textepsilon/, /u/, /o/, /\textopeno/, /a/. Each of which may be short or long, where the quantitative phonological contrast is based on differences in vowel length without additional qualitative differences.
Basaa is a tonal language, it has 4 tones: two register tones (high /H/ and low /L/) and two contours (rising /LH/ and falling /HL/). A low tone is orthographically unmarked while other tones are indicated by diacritics over a vowel or nasal sound (the marks varies in different orthographies).  There is the major tone rule – High Tone Spreading (HTS): /H-L/ sequence becomes [H-HL] on the surface (Hyman 2003). For example: /k\'{\textepsilon}mb{\textepsilon}/ -> [k\'{\textepsilon}mb\^{\textepsilon}] ‘goat’. Tones distinguish the meaning of words, so it is important to mark it in writing. For example:
\begin{quote}
\`M\`{\textepsilon} nl\`{\textopeno}  yani ‘I’ll come tomorrow’

\`M\`{\textepsilon} \`{n}l\^{\textopeno}  yani ‘I came yesterday’

\`M\`{\textepsilon}  nl\={\textopeno} yani ‘I'm gonna throw up tomorrow’

\`M\`{\textepsilon}  \`{n}l{\textopeno} yani ‘I threw up yesterday’
\end{quote}

\section{Related Literature}
\label{sec:length}

The task of tone prediction and orthographic conversion can be formalized as an 
Automatic Diacritic Restoration task and as a Spelling Correction task. Although the 
current generation of Basaa speakers uses the General Alphabet of Cameroon 
Languages, experimenting with tone prediction and orthographic conversion may be 
useful for further elaboration of NLP tools and techniques for Basaa.

\subsection{Automatic Diacritic Restoration Approaches}

There are many languages around the world that use diacritics. In recent years, NLP 
researchers have created many ASRS, e.g. for French \cite{15}, Croatian \cite{14}, Romanian \cite{16}, \cite{12}, \cite{10}, 
Spanish \cite{17}, \cite{3}, Maori \cite{2}, Vietnamese 
\cite{11}, and others. Among the low-resource African languages, we can 
highlight the works with the Igbo \cite{4} and Yoruba \cite{1}. 
Researchers divide approaches into two types - restoration of diacritics at the word or 
grapheme level (or at the letter level).
Researchers found out that many neural network architectures may be useful for that 
task: seq2seq, Residual Neural Networks, Convolutional Neural Networks. However, 
these approaches often require large datasets. The low number of resources available 
for the Basaa language imposes some restrictions on the corpuses that we can use. 
The task is complicated by the fact that before switching to the General Alphabet of 
Cameroon, Basaa had two types of orthography at once, which reduces the amount of 
unified data and forces us to make some special reservations regarding the input of the 
model.

\subsection{Sequence-Tagging Approach}

Sequence tagging is an NLP task where a class or label should be assigned to each 
token in a given input sequence. In this context, a word or a subword unit is referred to 
as a 'token' and receives a specific tag which means the transformations that the token 
must undergo. LaserTagger, a sequence-tagging model 
built on Tensorflow and BERT \cite{8} shows special effectiveness on 
small corpuses. On samples up to 10 thousand training examples, it significantly outperforms seq2seq. Another advantage of LaserTagger for tone prediction and 
orthographic conversion is that the training outcomes (tags) can be meaningfully 
interpreted.

\subsection{Seq2seq, T5 and mT5}

Sequence-to-sequence (seq2seq) algorithm was developed in 2014 \cite{19} and since 2018 was actively used by Google for machine translation. Its distinctive feature is that it turns one character’s sequence into another one using a RNN, LSTM or GRU where the context of each item is the output from the previous step. Since then, the algorithm has significantly expanded its scope, which now includes language translation, image captioning, conversational models and text summarization and any other tasks in which you need to deal with sequences of characters.
In 2020 Google presented a next-level algorithm, “Text-to-Text Transfer Transformer” (T5), an encoder-decoder model pre-trained on a masked language modeling “span-corruption” objective (the model is trained to reconstruct the masked-out tokens) \cite{20}. It converts all NLP problems into a text-to-text format. Another important distinguishing feature of the T5 is its scale: these models were pre-trained on around 1 trillion tokens of data and there are available from 60 million to 11 billion parameters. Later Google releases a multilingual version of T5, mT5, which allows you to work with any language, not just English \cite{23}.

\subsection{GECToR}

Grammatical Error Correction: Tag, Not Rewrite (GECToR) was presented in May 2020 \cite{22} and involves three distinct steps. The first step pre-processes the source sequence by tagging each token with a transformation tag, each transformation tag being added together to form the target sequence \cite{22}. An example of this can be seen with the input phrase, "A forty years old man go work." and the target phrase, "A forty-year-old man goes to work". "years" from the input phrase would recieve the KEEP tag as well as the MERGE HYPHEN tag for [years -> year -], and after "go" there would be an append tag for "to" which would look like "APPEND to". The second step involves reducing the Levenshtein distance (the amount of possible transitions the source sequence would need to go through in order to transform into the target sequence) \cite{22}. The third step requires conversion, where tokens which had been attributed more than one tag (such as to be kept, but also required to append a period at the end) would need to drop their KEEP tag in order to be left with the tags that would require transformation \cite{22}. The transformation would occur (if any was necessary), and thus the source sequence would become one iteration closer to becoming the target sequence. Looking at the previous example used, this would look like "years" losing its KEEP tag and keeping its MERG HYPHEN tag in order for the model to understand that a hyphen is needed to be added after it's transformation to "years". Each kept token, if not spelled exactly like the target text, is transformed into the target text used. For clarification, there are no transformation tags for adding missing morphemes or characters, as GECToR makes the necessary changes by changing the input word into the target word. The GECToR model runs through these three steps in as many iterations as needed until all necessary transformations have been made.

\section{Data}
Our dataset included a training set of 10,000 phrases, a validation set of 1,000 phrases, and a test set of 1,000 phrases.

\section{Model}
Earlier in the paper, we came to the conclusion that the GECToR model shows the SOTA results in solving our task. However, at the moment it exists an English version only. Moreover, its adaptation to other languages would require  a significant amount of linguistic information from dictionaries, as well as the use of a BERT pre-trained on a large Basaa corpus. It creates restrictions on the use of this approach for low-resource languages. Therefore, we used the second most popular approach: seq2seq, namely the mT5 model.

\section{Results}
First, we trained 7 seq2seq models based on mT5-small on un-preprocessed texts varying the number of epochs and the length of the sentence. In the case of seq2seq models, it is important to choose the optima for both hyperparameters. The best results were shown by the combination 7 epochs and sequence length equal to 40.

Later, we eliminated one-to-one correspondences between spellings and unified characters consisting of two characters, turning them into single-character characters. This also made it possible to reduce the dimensionality of the sequences.

\begin{table}
\centering
\begin{tabular}{lrl}
\hline \textbf{Parameters} & \textbf{CER} & \textbf{WER} \\ \hline
1 ep., length 73 & 46.047 & 119.65 \\
4 ep., length 73 & 26.825 & 69.709 \\
5 ep., length 25 & 33.480 & 64.535 \\
5 ep., length 35 & 26.989 & 64.436 \\
5 ep., length 60 & 26.541 & 76.234 \\
\textbf{7 ep., length 35} & \textbf{26.961} & \textbf{61.53} \\
10 ep., length 40 & 26.960 & 61.530 \\
10 ep., length 45 & 26.177 & 66.499 \\
\hline
\end{tabular}
\caption{\label{font-table} mT5 fine-tuning on un-processed texts}
\end{table}

The final model performed slightly better than the baseline, but this is an important result, given the special sensitivity of seq2seq models to the size of parallel corpora. We received CER equal to 12.6747 and WER equal to 40.1012 on pre-processed texts. The baseline CER for pre-processed texts is 14.1288 and WER is 40.8113.

\section{Conclusion}
In this paper we showed how a multilingual seq2seq model can be used for tone prediction and diacritic restoration on the example of Basaa. It allows to get results better than from a rule-based baseline but still depends on the quality of pre-processing and the size of parallel corpora.

\bibliography{acl2020}
\bibliographystyle{acl_natbib}


\end{document}